%% file: main.tex
\documentclass{article}
\usepackage{ijcai20}

\usepackage{times}
\usepackage{soul}
\usepackage{url}
\usepackage[hidelinks]{hyperref}
\usepackage[utf8]{inputenc}
\usepackage[small]{caption}
\usepackage{graphicx}
\usepackage{amsmath}
\usepackage{amssymb}
\usepackage{booktabs}
\usepackage{float}
\usepackage{caption}
\usepackage{color}
\usepackage{xcolor}
\usepackage{subcaption}
\usepackage{comment}
\usepackage{epstopdf}

\newcommand{\burak}[1]{ \color{red} [Burak: #1]  \color{black}}
\newcommand{\ayush}[1]{ \color{blue} [Ayush: #1]  \color{black}}

\title{Generating Interpretable Poverty Maps using Object Detection in Satellite Images}

\author{
Kumar Ayush$^1$\thanks{Equal Contribution}\and
Burak Uzkent$^1$\footnotemark[1]\and
Marshall Burke$^2$\and
David Lobell$^2$\and
Stefano Ermon$^1$
\\
\affiliations
$^1$Department of Computer Science, Stanford University\\
$^2$Department of Earth System Science, Stanford University\\
\{kayush, buzkent\}@cs.stanford.edu, mburke@stanford.edu, dlobell@stanford.edu, ermon@cs.stanford.edu
}

\begin{document}
\maketitle

\input{abstract.tex}

\input{introduction.tex}

\input{related_work.tex}

\input{datasets.tex}
\input{feature_extraction.tex}

\input{experiments.tex}

\input{conclusion.tex}
\bibliographystyle{unsrt}
\bibliography{ijcai20}
\input{appendix.tex}

\end{document}

%% file: abstract.tex
\begin{abstract}
Accurate  local-level  poverty  measurement is an  essential  task  for  governments and humanitarian organizations to track the progress towards improving livelihoods and distribute scarce resources. Recent computer vision advances in using satellite imagery to predict poverty have shown increasing accuracy, but they do not generate features that are interpretable to policymakers, inhibiting adoption by practitioners. Here we demonstrate an interpretable computational framework to accurately predict poverty at a local level by applying object detectors to high resolution (30cm) satellite images.  Using the weighted counts of objects as  features,  we  achieve  0.539 Pearson's $r^2$ in  predicting village level poverty in Uganda, a 31\% improvement over existing (and less interpretable) benchmarks. Feature importance and ablation analysis reveal intuitive relationships between object counts and poverty predictions. Our results suggest that interpretability does not have to come at the cost of performance, at least in this important domain.
\end{abstract}

%% file: introduction.tex
\section{Introduction}
Accurate measurements of poverty and related human livelihood outcomes critically shape the decisions of governments and humanitarian organizations around the world, and the eradication of poverty remains the first of the United Nations Sustainable Development Goals \cite{assembly2015sustainable}. However, reliable local-level measurements of economic well-being are rare in many parts of the developing world.  Such measurements are typically made with household surveys, which are expensive and time consuming to conduct across broad geographies, and as a result such surveys are conducted infrequently and on limited numbers of households.  For example, Uganda (our study country) is one of the best-surveyed countries in Africa, but surveys occur at best every few years, 
and when they do occur often only survey a few hundred villages across the whole country (Fig.~\ref{fig:pipeline}). Scaling up these ground-based surveys to cover more regions and more years would likely be prohibitively expensive for most countries in the developing world~\cite{jerven2017much}.  The resulting lack of frequent, reliable local-level information on economic livelihoods hampers the ability of governments and other organizations to target assistance to those who need it and to understand whether such assistance is having its intended effect.

To tackle this data gap, an alternative strategy has been to try to use passively-collected data from non-traditional sources to shed light on local-level economic outcomes.   Such work has shown promise in measuring certain indicators of economic livelihoods at local level. For instance, \cite{blumenstock2015predicting} show how features extracted from cell phone data can be used to predict asset wealth in Rwanda, and \cite{sheehan2019predicting} show how applying NLP techniques to Wikipedia articles can be used to predict asset wealth in multiple developing countries, and \cite{jean2016combining} show how a transfer learning approach that uses coarse information from nighttime satellite images to extract features from daytime high-resolution imagery can also predict asset wealth variation across multiple African countries.  

These existing approaches to using non-traditional data are promising, given that they are inexpensive and inherently scalable, but they face two main challenges that inhibit their broader adoption by policymakers. The first is the outcome being measured.  While measures of asset ownership are thought to be relevant metrics for understanding longer-run household well-being \cite{filmer2001estimating}, official measurement of poverty requires data on consumption expenditure (i.e. the value of all goods consumed by a household over a given period), and existing methods have either not been used to predict consumption data or perform much more poorly when predicting consumption than when predicting other livelihood indicators such as asset wealth \cite{jean2016combining}.  Second, interpretability of model predictions is key for whether policymakers will adopt machine-learning based approaches to livelihoods measurement, and current approaches attempt to maximize predictive performance rather than interpretability. This tradeoff, central to many problems at the interface of machine learning and policy \cite{murdoch2019definitions}, has yet to be navigated in the poverty domain.

Here we demonstrate an interpretable computational framework for predicting local-level consumption expenditure using object detection on high-resolution (30cm) daytime satellite imagery.  We focus on Uganda, a country with existing high-quality ground data on consumption where performance benchmark are available. We first train a satellite imagery object detector on a publicly available, global scale object detection dataset, called xView~\cite{lam2018xview}, which avoids location specific training and provides a more general object detection model. We then apply this detector to high resolution images taken over hundreds of villages across Uganda that were measured in an existing georeferenced household survey, and use extracted counts of detected objects as features in a final prediction of consumption expenditure. We show that not only does our approach substantially outperform previous performance benchmarks on the same task, it also yields features that are immediately and intuitively interpretable to the analyst or policy-maker.

%% file: related_work.tex
\section{Related Work}
\paragraph{Poverty Prediction from Imagery}
Multiple studies have sought to use various types of satellite imagery for local-level prediction of economic livelihoods. As already described, \cite{jean2016combining} train a CNN to extract features in  high-resolution daytime images using low-resolution nighttime images as labels, and then use the extracted features to predict asset wealth and consumption expenditure across five African countries. \cite{perez2017poverty} train a CNN to predict African asset wealth from lower-resolution (30m) multi-spectral satellite imagery, achieving similar performance to \cite{jean2016combining}. These approaches provide accurate methods for predicting local-level asset wealth, but the CNN-extracted features used to make predictions are not easily interpretable, and performance is substantially lower when predicting consumption expenditure rather than asset wealth. 

Two related papers use object detection approaches to predicting economic livelihoods from imagery.  \cite{gebru2017using} show how information on the make and count of cars detected in Google Streetview imagery can be used to predict socioeconomic outcomes at local level in the US.  This work is promising in a developed world context where streetview imagery is available, but challenging to employ in the developing world where such imagery is very rare, and where car ownership is uncommon.  In work perhaps closest to ours, an unpublished paper by \cite{engstrom2017poverty} use detected objects and textural features from high-resolution imagery to predict consumption in Sri Lanka, but model performance is not validated out of sample and the object detection approach is not described.

%% file: datasets.tex
\section{Problem Setup}
\subsection{Poverty Estimation from Remote Sensing Data}
The outcome of interest in this paper is consumption expenditure, which is the metric used to compute poverty statistics; a household or individual is said to be poor or in poverty if their measured consumption expenditure falls below a defined threshold (currently \$1.90 per capita per day). Throughout the paper we use ``poverty" as shorthand for ``consumption expenditure", although we emphasize that the former is computed from the latter. While typical household surveys measure consumption expenditure at the household level, publicly available data typically only release geo-coordinate information at the ``cluster" level -- which is a village in rural areas and a neighborhood in urban areas.  Efforts to predict poverty have thus focused on predicting at the cluster level (or more aggregated levels), and we do the same here. 
Let $\{(x_i, y_i, c_i)\}_{i=1}^N$ be a set of $N$ villages surveyed, where $c_i = (c^{lat}_i, c^{long}_i)$ is the latitude and longitude coordinates for cluster $i$, and $y_i \in \mathbb{R}$ is the corresponding average poverty index for a particular year. 

For each cluster $i$, we can acquire high resolution satellite imagery corresponding to the survey year $x_i \in \mathcal{I} = \mathbb{R}^{W \times H \times B}$, a $W \times H$ image with $B$ channels.
Following \cite{jean2016combining}, our goal is to learn a regressor $f:\mathcal{I} \rightarrow \mathbb{R}$ to predict the poverty index $y_i$ from $x_i$. Here our goal is to find a regressor that is both accurate and \textit{interpretable}, where we use the latter to mean a model that provides insight to a policy community on why it makes the predictions it does in a given location. 

\begin{table*}[!h]
\resizebox{0.95\linewidth}{!}{
\begin{tabular}{@{}llccccccccc@{}}
\toprule
\textbf{Fixed-Wing Aircraft} & \textbf{Passenger-Vehicle} & \textbf{Truck} & \textbf{Railway Vehicle} & \textbf{Maritime Vessel} & \textbf{Engineering Vehicle} & \textbf{Building} & \textbf{Helipad} & \textbf{Construction Site} & \textbf{Vehicle Lot} & \textbf{None} \\ \midrule
Small Aircraft & Small Car & Pickup Truck & Passenger Car & Motoboat & Tower Crane & Hut/ Tent &  &  &  & Pylon \\
Cargo & Bus & Utility Truck & Cargo Car & Sailboat & Container Crane & Shed &  &  &  & Shipping Container \\
 &  & Cargo Truck & Flat Car & Tugboat & Reach Stacker & Aircraft Hangar &  &  &  & Shipping Container Lot \\
 &  & Truck w/ Box & Tank Car & Barge & Straddle Carrier & Damaged Building &  &  &  & Storage Tank \\
 &  & Truck Tractor & Locomotive & Fishing Vessel & Mobile Crane & Facility &  &  &  & Tower Structure \\
 &  & Trailer &  & Ferry & Dump Truck &  &  &  &  & Helicopter \\
 &  & Truck w/ Flatbed &  & Yacht & Haul Truck &  &  &  &  &  \\
 &  & Truck w/ Liquid &  & Container Ship & Scraper/ Tractor &  &  &  &  &  \\
 &  &  &  & Oil Tanker & Front Loader &  &  &  &  &  \\
 &  &  &  &  & Excavator &  & \multicolumn{1}{l}{} & \multicolumn{1}{l}{} & \multicolumn{1}{l}{} & \multicolumn{1}{l}{} \\
 &  &  &  &  & Cement Mixer &  & \multicolumn{1}{l}{} & \multicolumn{1}{l}{} & \multicolumn{1}{l}{} & \multicolumn{1}{l}{} \\
 &  &  &  &  & Ground Grader &  & \multicolumn{1}{l}{} & \multicolumn{1}{l}{} & \multicolumn{1}{l}{} & \multicolumn{1}{l}{} \\
 &  &  &  &  & Crane Truck &  & \multicolumn{1}{l}{} & \multicolumn{1}{l}{} & \multicolumn{1}{l}{} & \multicolumn{1}{l}{} \\ \bottomrule
\end{tabular}}
\caption{Parent and child level classes in xView
. Originally, \emph{Helipad}, \emph{Construction Site}, and \emph{Vehicle Lot} are placed into the \emph{None} parent class. We change the structure slightly by using each one of them as an independent parent class. Finally, in our parent level detector we exclude the \emph{None} class resulting in 10 parent level classes. To train the child level detector, we use the original 60 child level classes including \emph{Helipad}, \emph{Construction Site}, and \emph{Vehicle Lot}.}
\label{tab:class_hierarchy}
\end{table*}
\subsection{Dataset}
\subsubsection{Socio-economic data}
\begin{figure*}[!h]
\centering
\includegraphics[width=\textwidth]{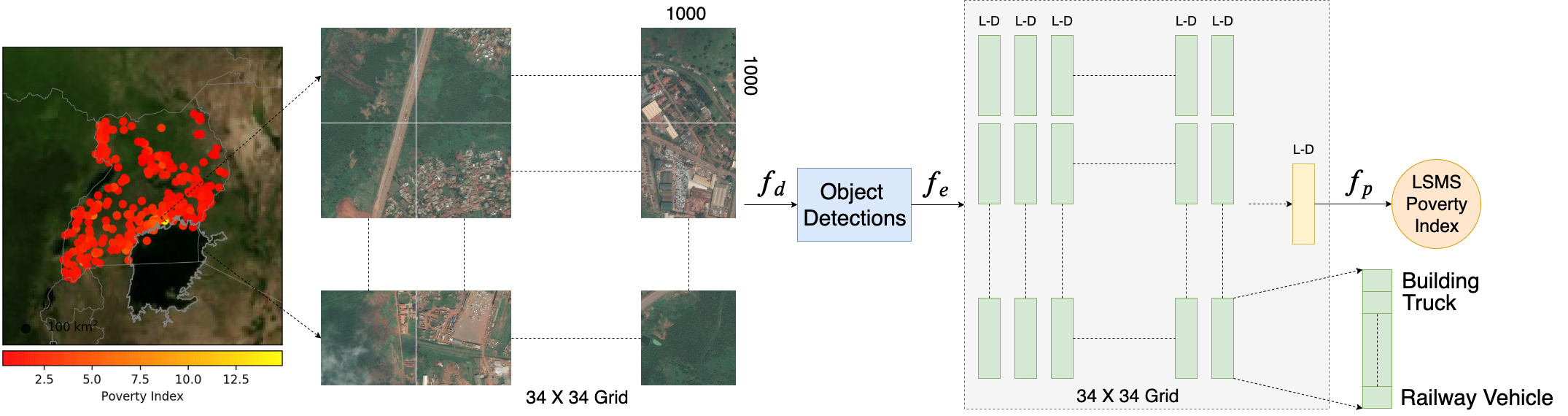}
\caption{Pipeline of the proposed approach.
For each cluster we acquire $1156$ images, arranged in a $34 \times 34$ grid, where each image is an RGB image of $1000 \times 1000$ px. 
We run an object detector on its $10 \times 10$ km$^2$ neighborhood and obtain the object counts for each xView class as a L-dimensional categorical feature vector. This is done by running the detector on every single image in a cluster, resulting in a $34 \times 34 \times L$ dimensional feature vector. Finally, we perform summation across the first two dimensions and get the feature vector representing the cluster, with each dimension containing the object counts corresponding to an object class. 
Given the cluster level feature vector, we regress the LSMS poverty score. 
}
\label{fig:pipeline}
\end{figure*}

The dataset comes from field Living Standards Measurement Study (LSMS) survey conducted in Uganda by the Uganda Bureau of Statistics between 2011 and 2012 \cite{lsms}. The LSMS survey we use here consists of data from 2,716 households in Uganda, which are grouped into unique locations called clusters. The
latitude and longitude location, $c_i = (c^{lat}_i, c^{long}_i)$,
of a cluster $i=\{1, 2, \dots, N\}$ is given, with noise of up to $5$ km added in each direction by the surveyers to protect privacy. Individual household locations in each cluster $i$ are also withheld to preserve anonymity. We use all $N = 320$ clusters in the survey to test the performance of our method in terms of predicting the average poverty index, $y_i$
for a group $i$. For each $c_i$,
the survey measures the poverty level by the per capital daily consumption in dollars. For simplicity, in this study, we name the per capital daily consumption in dollars as LSMS poverty score. We visualize the chosen locations on the map as well as their corresponding LSMS poverty scores in Fig. \ref{fig:pipeline}. From the figure, we can see that the surveyed locations are scattered near the border of states and high percentage of these locations have relatively low poverty scores.
\begin{table*}[!h]
\centering
\resizebox{0.85\linewidth}{!}{
\begin{tabular}{@{}lllllllllll@{}}
\toprule
\textbf{} & \textbf{Building} & \textbf{\begin{tabular}[c]{@{}l@{}}Fixed-Wing \\ Aircraft\end{tabular}} & \textbf{\begin{tabular}[c]{@{}l@{}}Passenger \\ Vehicle\end{tabular}} & \textbf{Truck} & \textbf{\begin{tabular}[c]{@{}l@{}}Railway\\ Vehicle\end{tabular}} &
\textbf{\begin{tabular}[c]{@{}l@{}}Maritime\\ Vessel\end{tabular}} &
\textbf{\begin{tabular}[c]{@{}l@{}}Engineering\\ Vehicle\end{tabular}} & \textbf{Helipad} & \textbf{\begin{tabular}[c]{@{}l@{}}Vehicle\\ Lot\end{tabular}} & \textbf{\begin{tabular}[c]{@{}l@{}}Construction\\ Site\end{tabular}} \\ \midrule
\textbf{AP} & 0.40 & 0.59 & 0.42 & 0.27 & 0.39 & 0.24 & 0.17 & 0.0 & 0.012 & 0.0003 \\
\textbf{AR} & 0.62 & 0.65 & 0.76 & 0.56 & 0.49 & 0.47 & 0.37 & 0.0 & 0.06 & 0.006 \\ \bottomrule
\end{tabular}}
  \caption{Class wise performance (average precision and recall) of YOLOv3 when trained using parent level classes (10 classes). 
  See \textbf{appendix} for the performance of YOLOv3 on child classes.
  }
  \label{table:parentClassLevelodresults}
\end{table*}
\subsubsection{Uganda Satellite Imagery}
\label{ugandaImages}
The satellite imagery, $x_i$
corresponding to cluster $c_i$
is represented by $K = 34 \times 34 = 1156$ images of $W=1000 \times H=1000$ pixels with $B=3$ channels, arranged in a $34 \times 34$ square grid. This corresponds to a $10$ km $\times$ $10$ km spatial neighborhood centered at $c_i$. We consider a large neighborhood to deal with the noise in the cluster coordinates. High resolution aerial images have been proven to be effective in many computer vision tasks including image recognition~\cite{uzkent2019learning,uzkent2019learningglobal}, object detection~\cite{lam2018xview,uzkent2019efficient}, and object tracking~\cite{uzkent2018tracking}. For this reason, we use the high resolution images from DigitalGlobe satellites with three bands (RGB) and 30cm pixel resolution. Figure~\ref{fig:pipeline} illustrates an example cluster from Uganda. Formally, we represent all the images corresponding to 
$c_i$ as a sequence of $K$ tiles as $x_i = \{x_i^j\}_{j=1}^K$.


\section{Fine-grained Detection on Satellite Images}
Contrary to existing methods for poverty mapping which perform end-to-end learning \cite{jean2016combining,sheehan2019predicting,perez2017poverty}, we use an intermediate object detection phase 
to first obtain 
interpretable features for subsequent poverty prediction.
However, we do not have object annotations for satellite images from Uganda. Therefore, we perform transfer learning by training an object detector on a different but related source dataset $\mathcal{D}^s$.

\subsection{Object Detection Dataset}
\label{xview_section}

We use xView \cite{lam2018xview}, as our source dataset. It is one of the largest and most diverse publicly available overhead imagery datasets for object detection. It covers over $1,400$ km$^2$ of the earth’s surface, with $60$ classes and approximately $1$ million labeled objects. The satellite images
are collected from DigitalGlobe satellites at $0.3$ m GSD, aligning with the GSD of our target region satellite imagery
$\{x_i\}_{i=1}^N$.
Moreover, xView uses a tree-structured ontology of 
classes. The classes are organized hierarchically similar to~\cite{deng2009imagenet,lai2011large} where children are more specific than their parents (e.g., \textit{fixed-wing aircraft} as a parent of \textit{small aircraft} and \textit{cargo plane}). 
Overall, there are $60$ child classes and $10$ parent classes. We show the hierarchy that we use in xView in Table~\ref{tab:class_hierarchy}.

%% file: feature_extraction.tex
\subsection{Training the Object Detector}

\paragraph{Models} Since we work on very large tiles ($\sim$3000$\times$3000 pixels), we only consider single stage detectors. Considering the trade off between run-time performance and accuracy on small objects, YOLOv3 \cite{redmon2018yolov3} outperforms other single stage detectors \cite{liu2016ssd,fu2017dssd} and performs almost on par with RetinaNet \cite{lin2017focal} but 3.8 $\times$ faster \cite{redmon2018yolov3} on small objects while running significantly faster than two-stage detectors~\cite{lin2017feature,shrivastava2016beyond}. Therefore, we use YOLOv3 object detector with a DarkNet53 \cite{redmon2018yolov3} backbone architecture.

\paragraph{Dataset Preparation} The xView dataset consists of $847$ large images (roughly $3000 \times 3000$ px). YOLOv3 is usually used with an input image size of $416 \times 416$ px. Therefore, we randomly chip $416\times 416$ px tiles from the xView images and discard tiles without any object of interest. This process results in $36996$ such tiles of which we use $30736$ tiles for training and $6260$ tiles for testing. 
\paragraph{Training and Evaluation}
We use the standard per-class average precision, mean average precision (mAP), and per-class recall, mean average recall (mAR) metrics \cite{redmon2018yolov3,lin2017focal} to evaluate our trained object detector. 
We fine-tune the weights of the YOLOv3 model, pre-trained on the ImageNet~\cite{deng2009imagenet}, using the training split of the xView dataset. 
Since xView has an ontology of parent and child level classes, we train two YOLOv3 object detectors using parent level and child level classes seperately. 
 
After training the models, we validate their performance on the test set of xView. The detector trained using parent level classes (10 classes) achieves mAP of 0.248 and mAR of 0.42. On the other hand, the one trained on child classes achieves mAP of 0.0745 and mAR of 0.242.
Table~\ref{table:parentClassLevelodresults} shows the class-wise performance of the parent-level object detector on the test set. For comparison, \citeauthor{lam2018xview} report 0.14 mAP, but they use a separate validation and test set in addition to the training set (which are not publicly available) so the models are not directly comparable.
While not state of the art, our detector reliably identifies objects, especially at the parent level. 
\subsection{Object Detection on Uganda Satellite Images}
\label{fd}



As described in Section \ref{ugandaImages}, each $x_i$ is represented by a set of $K$ images, $\{x_i^j\}_{j=1}^K$.
Each $1000 \times 1000$ px tile (i.e. $x_i^j$) is further chipped into $9$ $416\times 416$ px small tiles (with overlap of 124 px) and fed to YOLOv3.

Although the presence of objects across tile borders could decrease performance, this method is highly parallelizable and enables us to scale to very large regions. 
We perform object detection on $320 \times 1156 \times 9$ chips (more than $3$ million images), which takes about a day and a half using 4 NVIDIA 1080Ti GPUs. In total, we detect 768404 objects. 
Each detection
is denoted by a tuple $(x_c, y_c, w, h, l, s)$, where $x_c$ and $y_c$ represent the center coordinates of the bounding box, $w$ and $h$ represent the width and height of the bounding box, $l$ and $s$ represent the object class label and class confidence score. In Section \ref{featureSection}, we explain how we use these details to create interpretable features. Additionally, we experiment with object detections obtained at different confidence thresholds which we discuss in Section \ref{povertyexperiments}.

\paragraph{Transfer performance in Uganda} 

\begin{figure}[!h]
    \centering
    \includegraphics[width=0.23\textwidth]{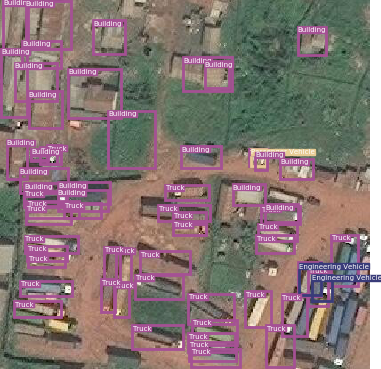}
    \hspace{0.3em}
    \includegraphics[width=0.22\textwidth]{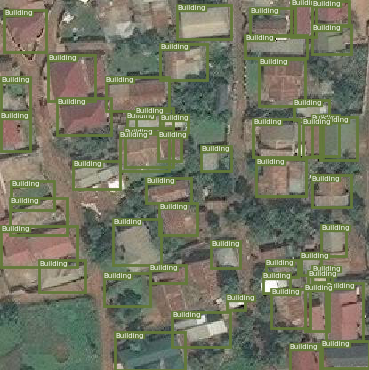}
    \caption{Sample detection results from Uganda. Zoom-in is recommended to visualize the bounding box classes. See \textbf{appendix} for more examples.
    }
    \label{fig:detect}
\end{figure}

The absence of ground truth object annotations for our Uganda imagery $\{x_i^j\}_{j=1}^K$ prevents us from quantitatively measuring the detector's performance on Uganda satellite imagery. However, we manually annotated 10 images from the Uganda dataset together with the detected bounding boxes to measure the detector's performance on building and truck classes. We found that the detector achieves about $50\%$, and $45\%$ AR for Building and Truck which is slightly lower than the AR scores for the same classes on the xView test set. We attribute this slight difference to the problem of domain shift and we plan to address this problem via domain adaptation in a future work. To qualitatively test the robustness of our xView-trained object detector, we also visualize its performance on two representative tiles in Fig. \ref{fig:detect}. The detection results prove the effectiveness of transferring the YOLOv3 model to DigitalGlobe imagery it has not been trained on. 
\section{Fine-level Poverty Mapping}
\subsection{Feature Extraction from Clusters}
\label{featureSection}
Our object detection pipeline outputs 
$z_i = \{z_i^j\}_{k=1}^K$, which consists of $K$ sets and each set $z_i^j \in z_i$ consists of 
$n_i^j$ object detections for each tile $x_i^j$ of $x_i$. We use the $n_i^j$ object detections to generate a $L$-dimensional vector, $\mathbf{v}_i^j \in \mathbb{R}^L$ (where $L$ is the number of object labels/classes), by counting the number of detected objects in each class with each object weighted by its confidence score or size or their combination (details below). This process results in $K$ $L$-dimensional vectors $\mathbf{v}_i = \{\mathbf{v}_i^j\}_{k=1}^K$. Finally, we aggregate these $K$ vectors into a single $L$-dimensional categorical feature vector $\mathbf{m}_i$ by summing over tiles: $
    \mathbf{m}_i = \sum_{j=1}^{K} \mathbf{v}_i^j
$. While many other options are possible, in this work we explore four types of features:

\paragraph{Counts} \textit{Raw object counts corresponding to each class}. 
As mentioned earlier, we collapse the $n_{i}^{j}$ object detections corresponding to $z^{j}_{i}$ into a $L$-dimensional categorical feature vector $\mathbf{v}_{i}^{j}$. Here, each dimension represents an object class and contains the number of objects detected corresponding to that class. We aggregate these $K$ $L$-dimensional vectors ($\mathbf{v}_{i}^{j}$) into a single $L$-dimensional categorical feature vector $\mathbf{m}_t^i \in \mathcal{M}_t$. 
For each class $\ell \in \{1, 2, \dots, L\}$


\begin{equation}
    \mathbf{v}_i^j[\ell] = \sum_{k=1}^{n_i^j} 1*\textbf{1}[\ell == o_k[l]]
\end{equation}

\paragraph{Confidence$\times$Counts} \textit{Each detected object is weighted by its class confidence score}. The intuition is to reduce the contributions of less confident detections. 
Here each dimension corresponds to the sum of class confidence scores of the detected objects of that class. 
For each class $\ell \in \{1, 2, \dots, L\}$

\begin{equation}
    \mathbf{v}_i^j[\ell] = \sum_{k=1}^{n_i^j} o_k[s]*\textbf{1}[\ell == o_k[l]]
\end{equation}
\paragraph{Size$\times$Counts} \textit{Each detected object is weighted by its bounding box area}. We posit that weighting based on area coverage of an object class can be an important factor. For example, an area with $10$ big buildings might have a different wealth level than an area with $10$ small buildings. Each dimension in $\mathbf{m}_i$ contains the sum of areas of the bounding boxes of the detected objects of that class. 
For each class $\ell \in \{1, 2, \dots, L\}$
\begin{equation}
    \mathbf{v}_i^j[\ell] = \sum_{k=1}^{n_i^j} o_k[w]*o_k[h]*\textbf{1}[\ell == o_k[l]]
\end{equation}



\paragraph{(Confidence, Size)$\times$Counts} \textit{Each detected object is weighted by its class confidence score and the area of its bounding box}.
We concatenate the \textit{Confidence} and \textit{Size} based features to create a $2L$-dimensional vector.

\subsection{Models, Training and Evaluation} 
Given the cluster level categorical feature vector,  $\mathbf{m}_i$, we estimate its poverty index, $y_i$ with a regression model. 
Since we value interpretability, we consider Gradient Boosting Decision Trees, Linear Regression, Ridge Regression, and Lasso Regression.
As we regress directly on the LSMS poverty index, we quantify the performance of our model using the square of the Pearson correlation coefficient (Pearson’s $r^2$). Pearson's $r^2$, provides a measure of how well observed outcomes are replicated by the model. 
This metric was chosen so that comparative analysis could be performed with previous literature \cite{jean2016combining}. Pearson’s $r^2$ is invariant under separate changes in scale between the two variables. This allows the metric to provide insight into the ability of the model to distinguish between poverty levels. This is relevant for many downstream poverty tasks, including the distribution of program aid under a fixed budget (where aid is disbursed to households starting with the poorest, until the budget is exhausted), or in the evaluation of anti-poverty programs, where outcomes are often measured in terms of percentage changes in the poverty metric.
Due to small size of the dataset, we use a Leave-one-out cross validation (LOOCV) strategy. 
Since nearby clusters could have some geographic overlap, we remove clusters which are overlapping with the test cluster from the train split to avoid leaking information to the test point.

%% file: experiments.tex
\section{Experiments}
\label{experiments}

\subsection{Poverty Mapping Results}
\label{povertyexperiments}
\begin{figure}[!h]
     \centering
     \begin{subfigure}[b]{0.23\textwidth}
         \centering
         \includegraphics[width=\textwidth]{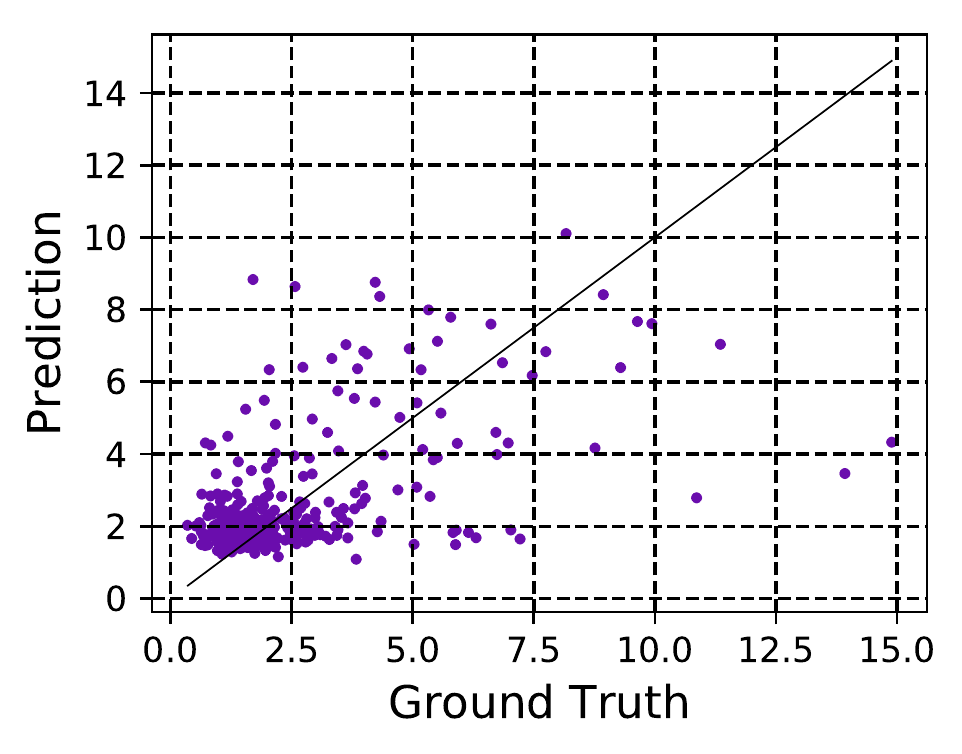}
         \caption{Counts}
         \label{fig:num}
     \end{subfigure}
     \begin{subfigure}[b]{0.23\textwidth}
         \centering
         \includegraphics[width=\textwidth]{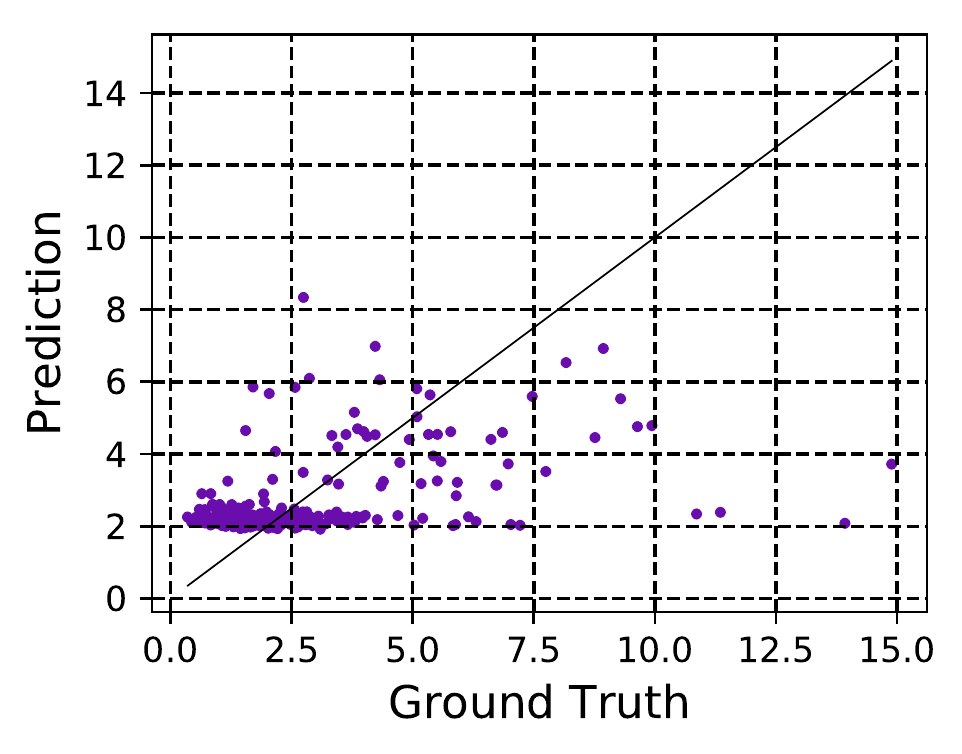}
         \caption{Confidence $\times$ Counts}
         \label{fig:score}
     \end{subfigure}
     \begin{subfigure}[b]{0.23\textwidth}
         \centering
         \includegraphics[width=\textwidth]{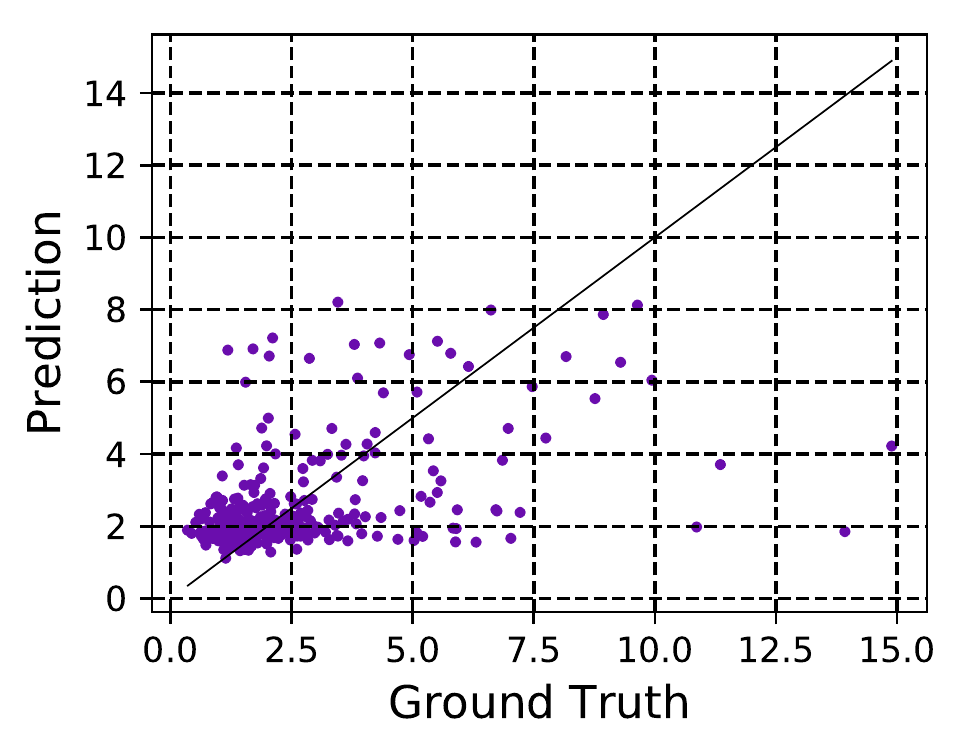}
         \caption{Size $\times$ Counts}
         \label{fig:size}
     \end{subfigure}
     \begin{subfigure}[b]{0.23\textwidth}
         \centering
         \includegraphics[width=\textwidth]{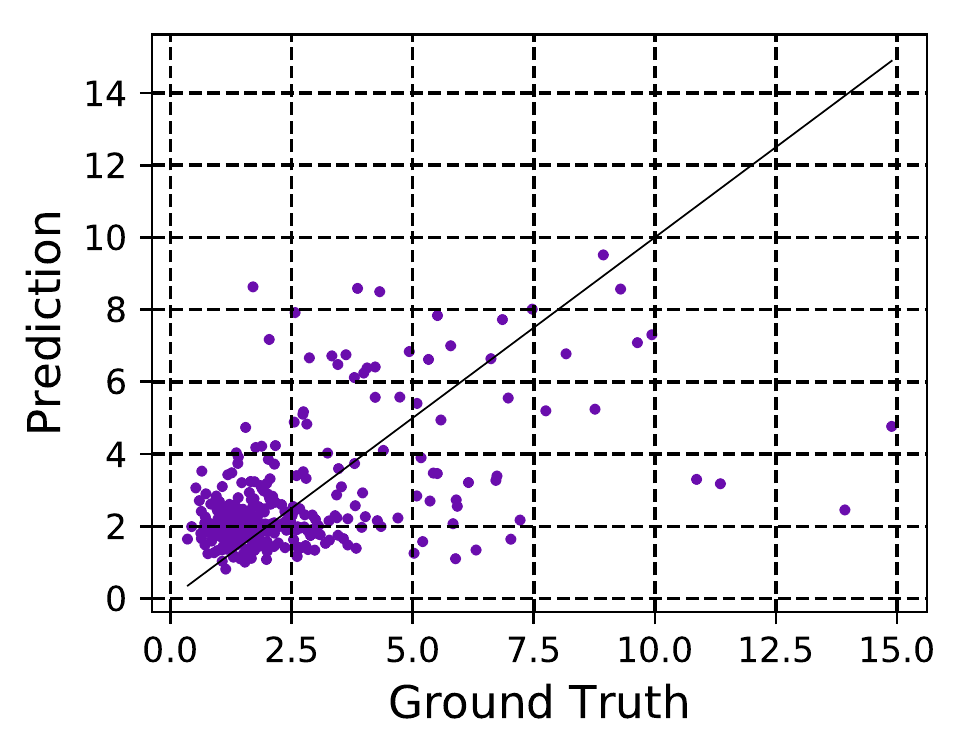}
         \caption{(Conf., Size) $\times$ Counts}
         \label{fig:scoresize}
     \end{subfigure}
        \caption{Regression result of GBDT using parent level counts.
        }
        \label{fig:gbdt}
\end{figure}

\textbf{Quantitative Analysis} Table \ref{table:rsqboth} shows the results of LSMS poverty prediction in Uganda. The object detections are obtained using a 0.6 confidence threshold (the effect of this hyper-parameter is evaluated below). The best result of $0.539$ Pearson's $r^2$ is obtained using GBDT trained on parent level \textit{Raw object Counts} features (red color entry). 
A scatter plot of GBDT predictions v.s. ground truth is shown in Fig.~\ref{fig:gbdt}. It can be seen that our GBDT model can explain a large fraction of the variance in terms of object counts automatically identified in high resolution satellite images.  To the best of our knowledge, this is the first time this capability has been demonstrated with a rigorous and reproducibile out-of-sample evaluation (see however the related but unpublished paper by \citeauthor{engstrom2017poverty}).

We observe that GBDT performs consistently better than other regression models across the four features we consider. 
As seen in Table~\ref{table:rsqboth}, object detection based features deliver positive $r^2$ with a simple linear regression method which suggests that they have positive correlation with LSMS poverty scores. 
However, the main drawback of linear regression against GBDT is that it predicts negative values, which is not reasonable as poverty indices are non-negative. In general, the features are useful, but powerful regression models are still required to achieve better performance.

We also find that finer-grained object detections (at the child level in the xView class hierarchy) can perform better than the coarser ones (second and third best) in some cases. This is likely because although they convey more information, detection and classification is harder at the finer level (see performance drop in Section 4), likely resulting in noisier predictions.
Additionally, parent level features are more suited for interpretability, due to household level descriptions, which we show later.

\begin{table}[!h]
\small
\textcolor{red}{\textbf{---}} Best \hspace{2em} \textcolor{blue}{\textbf{---}} Second Best \hspace{2em}
\textcolor{green}{\textbf{---}} Third Best
\centering
\resizebox{0.99\columnwidth}{!}{
\begin{tabular}{@{}ccccccccc@{}}
\toprule
\textbf{Features/Method} & \multicolumn{2}{c}{\textbf{GBDT}} & \multicolumn{2}{c}{\textbf{Linear}} & \multicolumn{2}{c}{\textbf{Lasso}} & \multicolumn{2}{c}{\textbf{Ridge}} \\ \midrule
 & \textbf{Parent} & \textbf{Child} & \textbf{Parent} & \textbf{Child} & \textbf{Parent} & \textbf{Child} & \textbf{Parent} & \textbf{Child} \\
\textbf{Counts} & \textcolor{red}{0.539} & 0.508 & 0.311 & 0.324 & 0.312 & 0.46 & 0.311 & 0.329 \\
\textbf{Confidence $\times$ Counts} & 0.466 & 0.485 & 0.305 & 0.398    & 0.305 & 0.461 & 0.305 & 0.409 \\
\textbf{Size $\times$ Counts} & 0.455 & \textcolor{blue}{0.535} & 0.363 & 0.47      & 0.363 & 0.476 & 0.363 & 0.47 \\
\textbf{(Conf., Size) $\times$ Counts} & 0.495 & \textcolor{green}{0.516} & 0.411 & 0.369 & 0.418 & 0.343 & 0.411 & 0.476 \\ \bottomrule
\end{tabular}}
  \caption{LSMS poverty score prediction results in Pearson's $r^2$ using parent level features (YOLOv3 trained on $10$ classes) and child level features (YOLOv3 trained on $60$ classes).}
  \label{table:rsqboth}
\end{table}


\paragraph{Comparison to Baselines and State-of-the-Art} 
We compare our method with two baselines and a state-of-the-art method: (a) \textbf{NL-CNN} where we regress the LSMS poverty scores using a 2-layer CNN with Nightlight Images ($48 \times 48$ px) representing the clusters in Uganda as input, (b) \textbf{RGB-CNN} where we regress the LSMS poverty scores using ImageNet~\cite{deng2009imagenet} pretrained ResNet-18 \cite{he2016deep} model with central tile representing $c_i$ as input, and (c) \textbf{Transfer Learning with Nightlights}, \cite{jean2016combining} proposed a transfer learning approach where nighttime light intensities are used as a data-rich proxy.

Results are shown in  Table~\ref{table:baseline}. Our model substantially outperforms all three baselines, including published state-of-the-art results on the same task in \cite{jean2016combining}.  We similarly outperform the NL-CNN baseline, a simpler version of which (scalar nightlights) is often used for impact evaluation in policy work \cite{donaldson2016view}.  Finally, the performance of the RGB-CNN baseline reveals the limitation of directly regressing CNNs on daytime images, at least in our setting with small numbers of labels. As discussed below, these performance improvements do not come at the cost of interpretability -- rather, our model predictions are much more interpretable than each of these three baselines. 



\begin{table}[!h]
\centering
\resizebox{0.85\columnwidth}{!}{
\begin{tabular}{@{}lllll@{}}
\toprule
\textbf{Method} & \textbf{RGB-CNN} & \textbf{NL-CNN} & \textbf{\cite{jean2016combining}} & \textbf{Ours} \\ \midrule
\textbf{$r^2$} & 0.04 & 0.39 & 0.41 & \textbf{0.54} \\
\bottomrule
\end{tabular}}
  \caption{Comparison with baseline and state-of-the-art methods.}
  \label{table:baseline}
\end{table}
\begin{figure}[!b]
    \centering
    \includegraphics[width=0.24\textwidth]{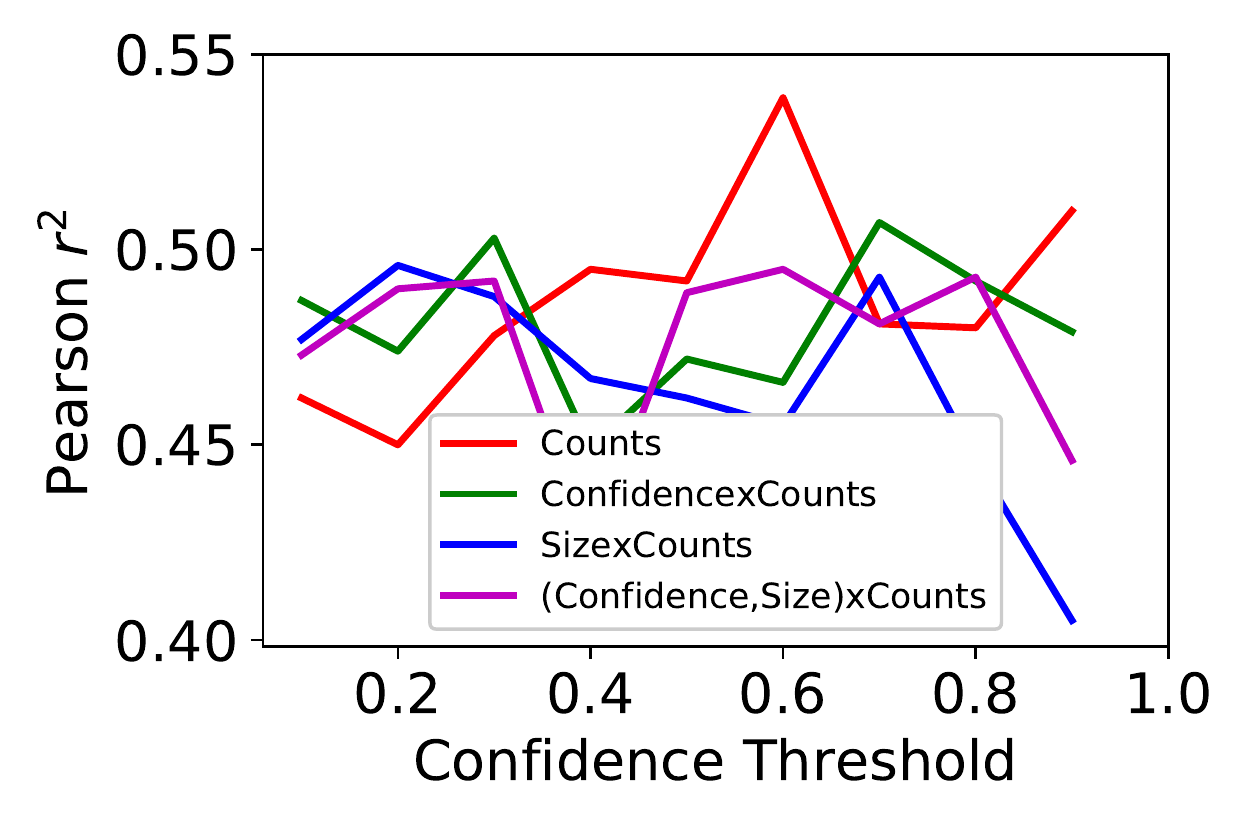}
    \includegraphics[width=0.235\textwidth]{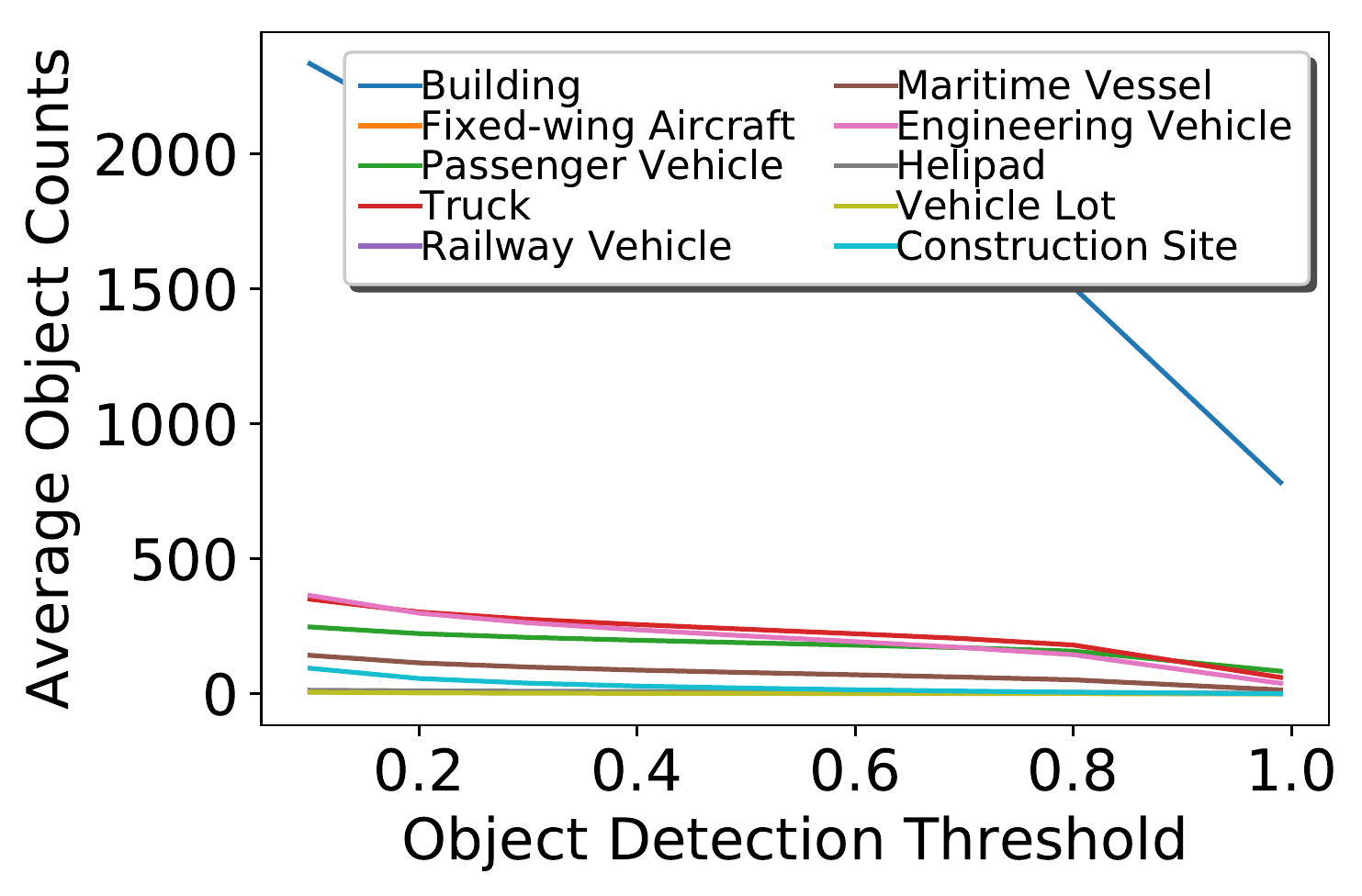}
    \caption{\textbf{Left:} Poverty score regression results of GBDT model using object detection features (parent level class features) at different confidence thresholds. \textbf{Right:} Average object counts across clusters for each parent class at difference confidence thresholds.
    }
    \label{fig:results_vs_threshold}
\end{figure}

\textbf{Impact of Detector's Confidence Threshold} Finally, we analyze the effect of confidence threshold for object detector on the poverty prediction task in Fig.~\ref{fig:results_vs_threshold}. We observe that when considering only \emph{Counts} features, we get the best performance at 0.6 threshold. However, even for very small thresholds, we achieve around $0.3$-$0.5$ Pearson's $r^{2}$ scores. We explore this finding in Fig. 3b, and observe that the \emph{ratio of classes in terms of number of bounding boxes remain similar} across different thresholds. These results imply that the ratio of object counts is perhaps more useful than simply the counts themselves -- an insight also consistent with the substantial performance boost from GBT over unregularized and regularized linear models in Table 1. 

\subsection{Interpretability}
\label{interpret}
Existing approaches to poverty prediction using unstructured data from satellites or other sources have understandably sought to maximize predictive performance \cite{jean2016combining,perez2017poverty,sheehan2019predicting}, but this has come at the cost of interpretability, as most of the extracted features used for prediction do not have obvious semantic meaning. While (to our knowledge) no quantitative data have been collected on the topic, our personal experience on multiple continents over many years is that the lack of interpretability of CNN-based poverty predictions can make policymakers understandably reluctant to trust these predictions and to use them in decision-making.  Enhancing the interpretability of ML-based approaches more broadly is thought to be critical component of successful application in many policy domains \cite{doshi2017towards}.

Relative to an end-to-end deep learning approach, our two-step approach with object detection provides categorical features that are easily understood.  We now explore whether these features also have an intuitive mapping to poverty outcomes in three analyses.  
\paragraph{Explanations via SHAP} In this section, we explain the effect of individual features (parent level GBDT model) on poverty score predictions using SHAP (SHapley Additive exPlanations) \cite{NIPS2017_7062}. SHAP is a game theoretic approach to explain the output of any machine learning model. It connects optimal credit allocation with local explanations using the classic Shapley values from game theory and their related extensions.
We particularly use TreeSHAP \cite{lundberg2018consistent} which is a variant of SHAP for tree-based machine learning models.
TreeSHAP significantly improves the interpretability of tree-based models through a) a polynomial time algorithm to compute optimal explanations based on game theory, b) explanations that directly measure local feature interaction effects, and c) tools for understanding global model structure based on combining many local explanations of each prediction.

\begin{figure}[!h]
    \centering
    \includegraphics[width=0.265\textwidth]{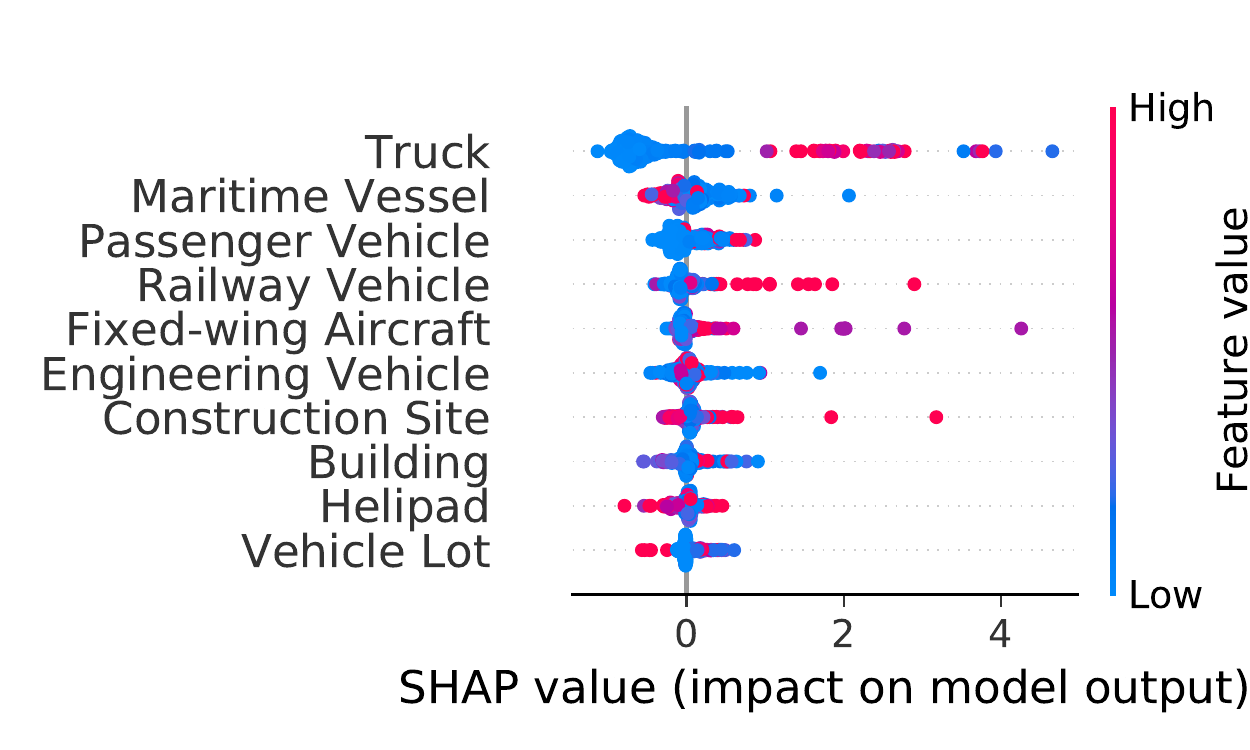}
    \label{fig:featureImportance}
    \hspace{0.3em}
    \includegraphics[width=0.195\textwidth]{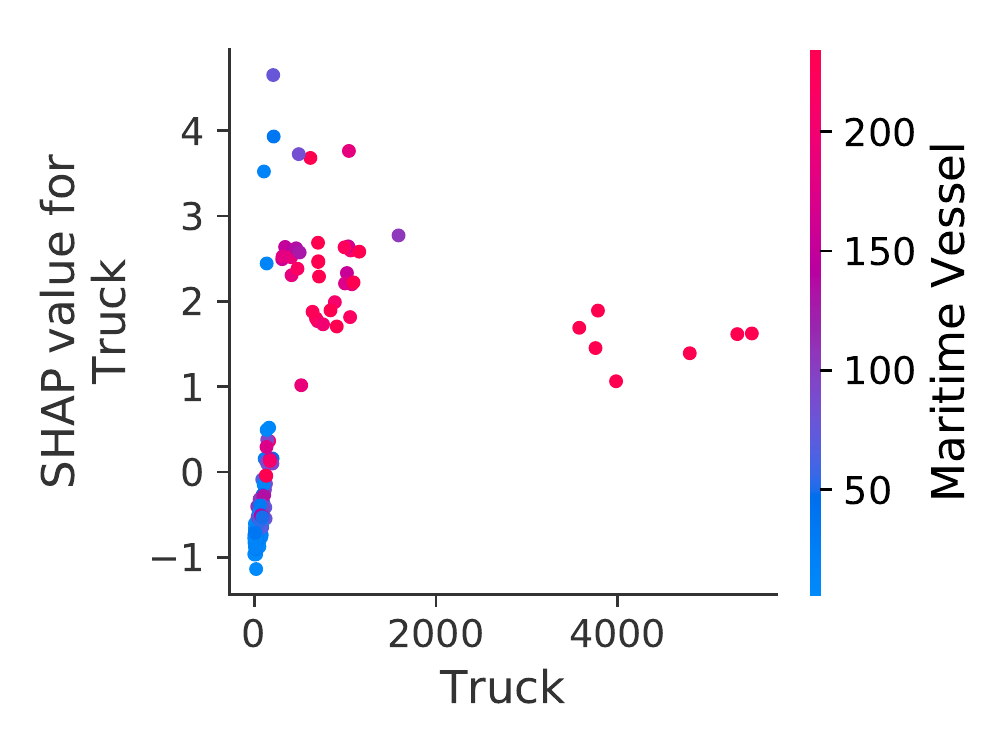}
    \caption{\textbf{Left}: Summary of the effects of all the features. \textbf{Right}: Dependence plot showing the effect of a single feature across the whole dataset. In both figures, the color represents the feature value (red is high, blue is low). See \textbf{appendix} for further dependence plots.
    }
    \label{fig:shap}
\end{figure}

To get an overview of which features are most important for a model we plot the SHAP values of every feature for every sample. The plot in Figure~\ref{fig:shap} (left) sorts features by the sum of SHAP value magnitudes over all samples, and uses SHAP values to show the distribution of the impacts each feature has on the model output. The color represents the feature value (red high, blue low). We find that \textit{Truck} tends to have a high impact on the model's output. Higher \#\textit{Trucks} pushes the output to a higher value and low \#\textit{Trucks} has a negative impact on the output, thereby lowering the predicted value.
\begin{figure}[!h]
    \centering
    \includegraphics[width=0.225\textwidth]{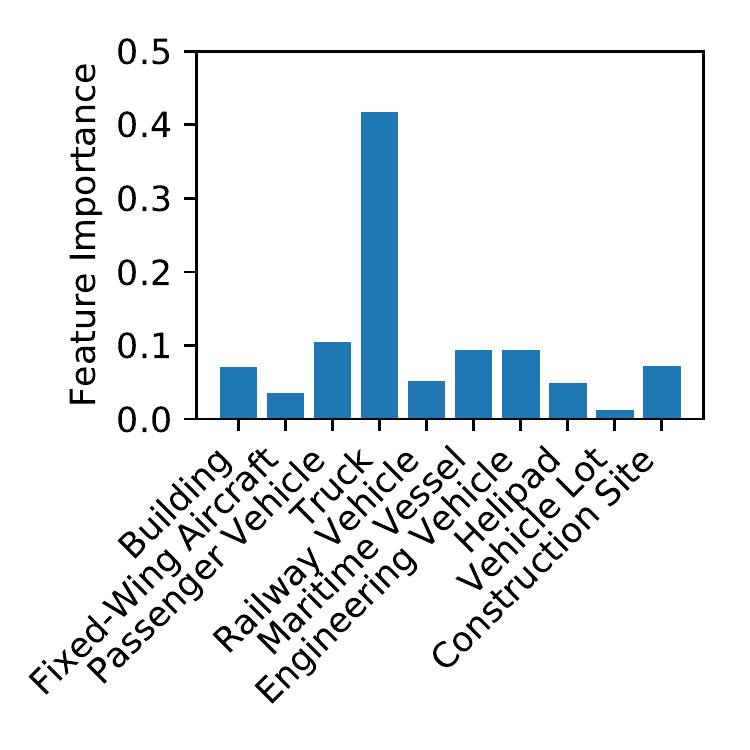}
    \label{fig:featureImportance}
    \includegraphics[width=0.225\textwidth]{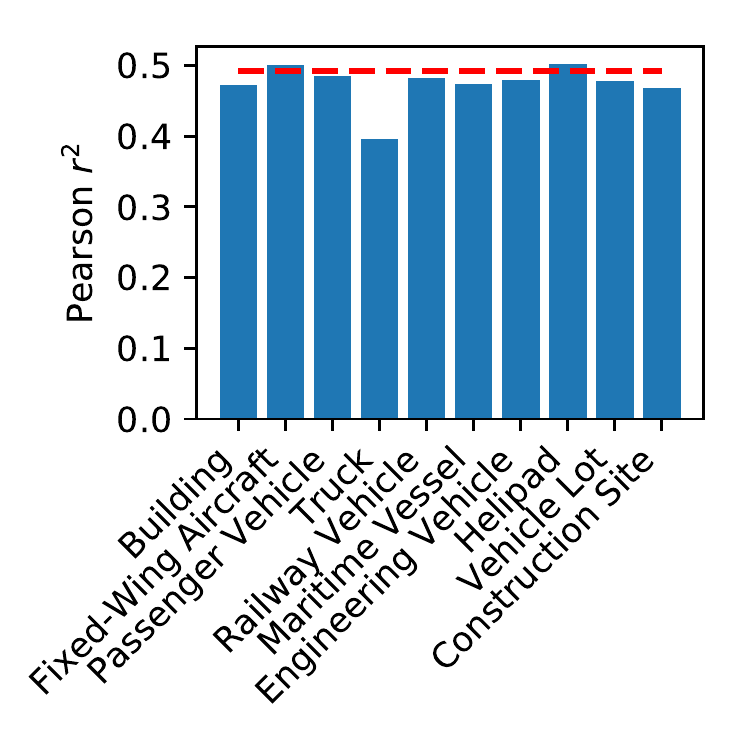}
    \caption{\textbf{Left}: Feature Importance of parent classes in the GBDT model. 
    \textbf{Right}: Ablation analysis where the red line represents the GBDT's performance when including all the parent classes. 
    }
    \label{fig:ablation}
\end{figure}

To understand how the Truck feature effects the output of the model we plot the SHAP value of \textit{Truck} feature vs. the value of the Truck feature for all the examples in the dataset. Since SHAP values represent a feature's responsibility for a change in the model output, the plot in Figure \ref{fig:shap} (right) represents the change in predicted poverty score as \textit{Truck} feature changes and also reveals the interaction between \textit{Truck} feature and \textit{Maritime Vessel} feature. 
We find that for small \#\textit{Trucks}, low \#\textit{Maritime Vessels} decreases the \textit{Truck} SHAP value. This can be seen from the set of points that form a vertical line (towards bottom left) where the color changes from blue (low \#\textit{Maritime Vessels}) to red (high \#\textit{Maritime Vessels}) as Truck SHAP value increases.


\paragraph{Feature Importance} 
We also plot the sum of SHAP value magnitudes over all samples for the various features (feature importance).
Figure~\ref{fig:ablation} (left) shows the importance of the 10 features (parent level features) in poverty prediction. \textit{Truck} has the highest importance. It is followed by \emph{Passenger Vehicle}, \emph{Maritime Vessel}, and \emph{Engg. Vehicle} with similar feature importances.

\paragraph{Ablation Analysis} 
Finally, we run an ablation study by training the regression model using all the categorical features in the train set and at test time we eliminate a particular feature by collapsing it to zero. We perform this ablation study with the parent level features as it provides better interpretability. Consistent with the feature importance scores, in Figure~\ref{fig:ablation} we find that when \textit{Truck} feature is eliminated at test time, the Pearson's $r^2$ value is impacted most.

%% file: conclusion.tex
\section{Conclusion}
In this work, we attempt to predict consumption expenditure from high resolution satellite images. We propose an efficient, explainable, and transferable method that combines object detection and regression. This model achieves a Pearson's $r^2$ of $0.54$ in predicting village level consumption expenditure in Uganda, even when the provided locations are affected by noise (for privacy reasons) and the overall number of labels is small ($\sim$300). The presence of trucks appears to be particularly useful for measuring local scale poverty in our setting. We also demonstrate that our features achieve positive results even with simple linear regression model. Our results offer a promising approach for generating interpretable poverty predictions for important livelihood outcomes, even in settings with limited training data. 

%% file: appendix.tex
\appendix
\section{Additional Results}
\begin{figure}[!h]
    \centering
    \includegraphics[width=0.23\textwidth]{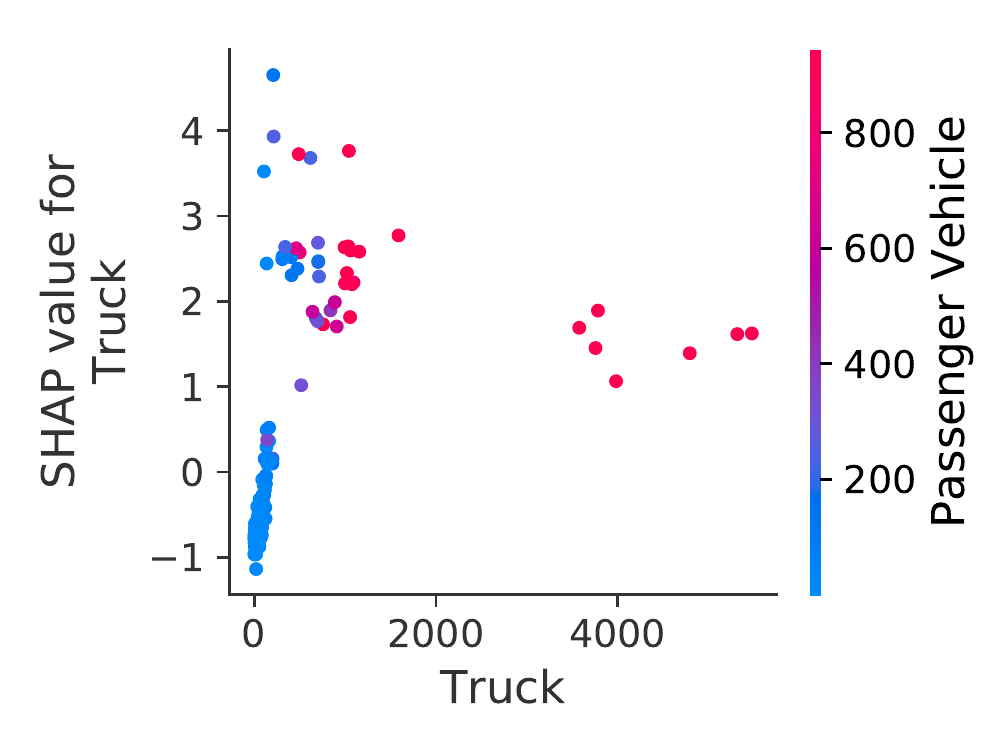}
    \hspace{0.4em}
    \includegraphics[width=0.23\textwidth]{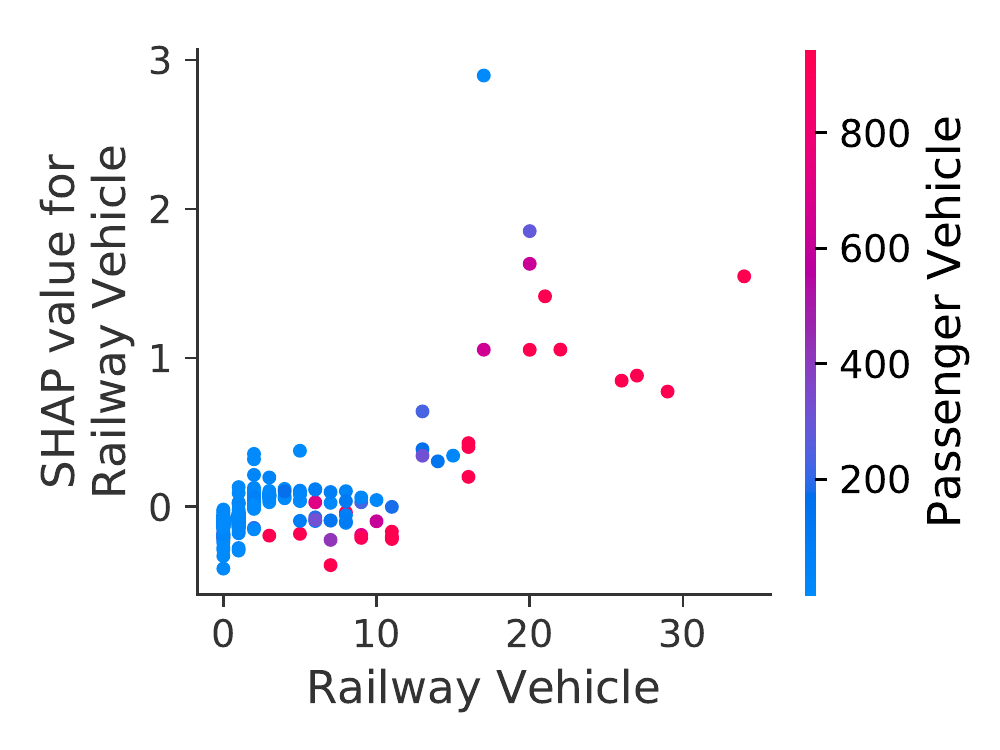}
    \centering
    \includegraphics[width=0.23\textwidth]{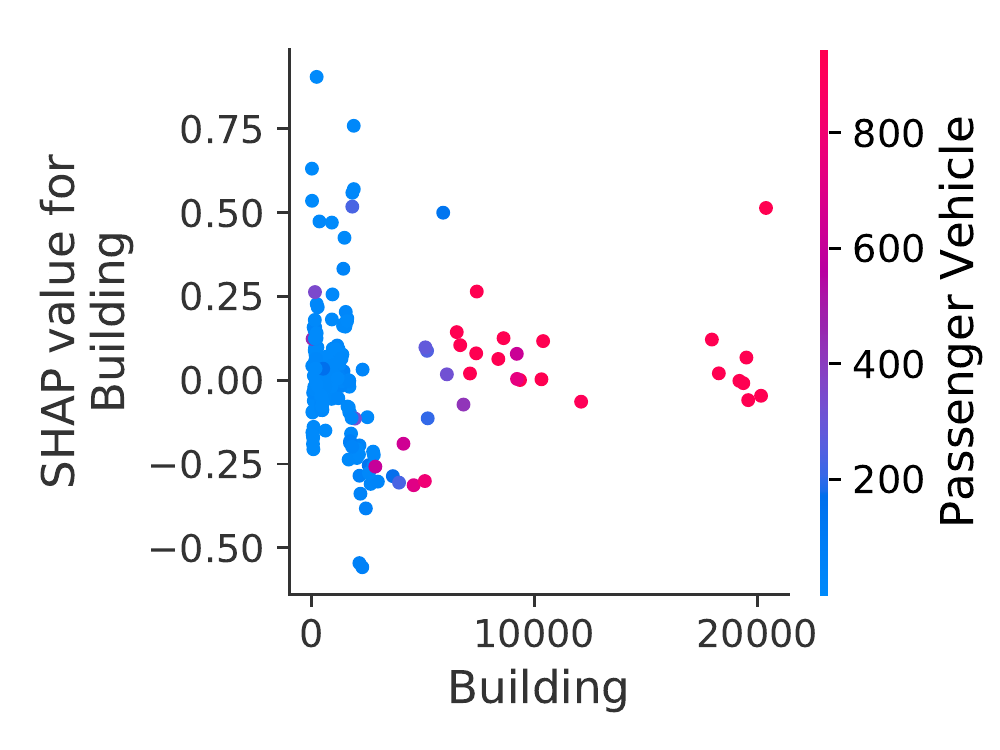}
    \hspace{0.4em}
    \includegraphics[width=0.23\textwidth]{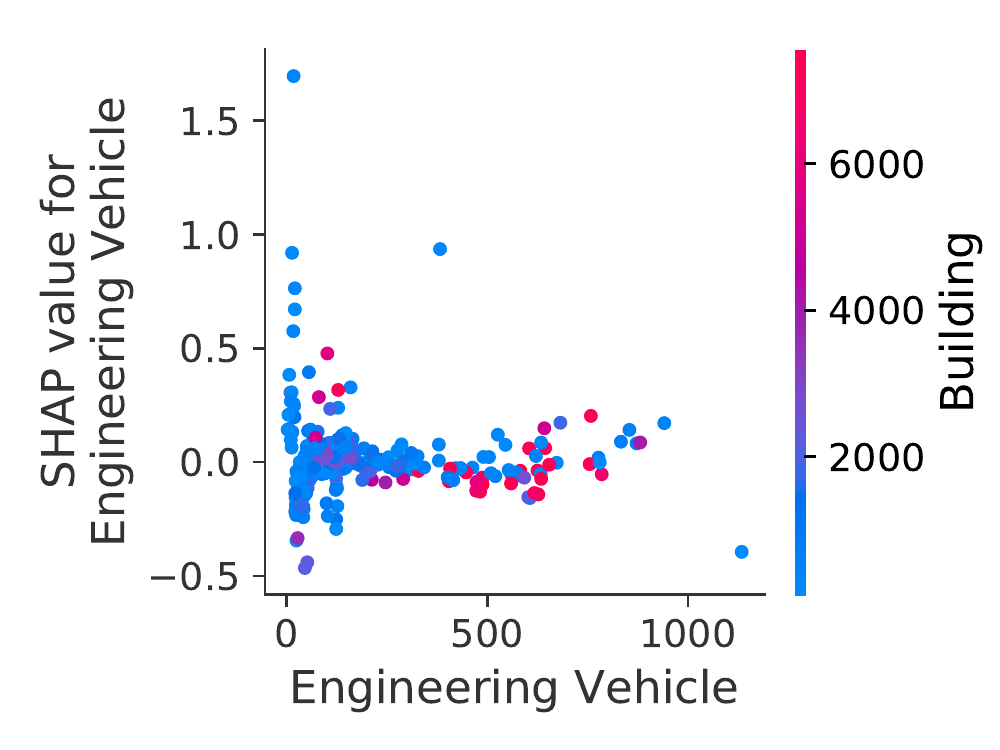}
    \caption{Dependence plots showing the effect of a single feature across the whole dataset. In both figures, the color represents the feature value (red is high, blue is low)}
    \label{fig:shap1}
\end{figure}
\begin{figure}[!t]
    \centering
    \includegraphics[width=0.23\textwidth]{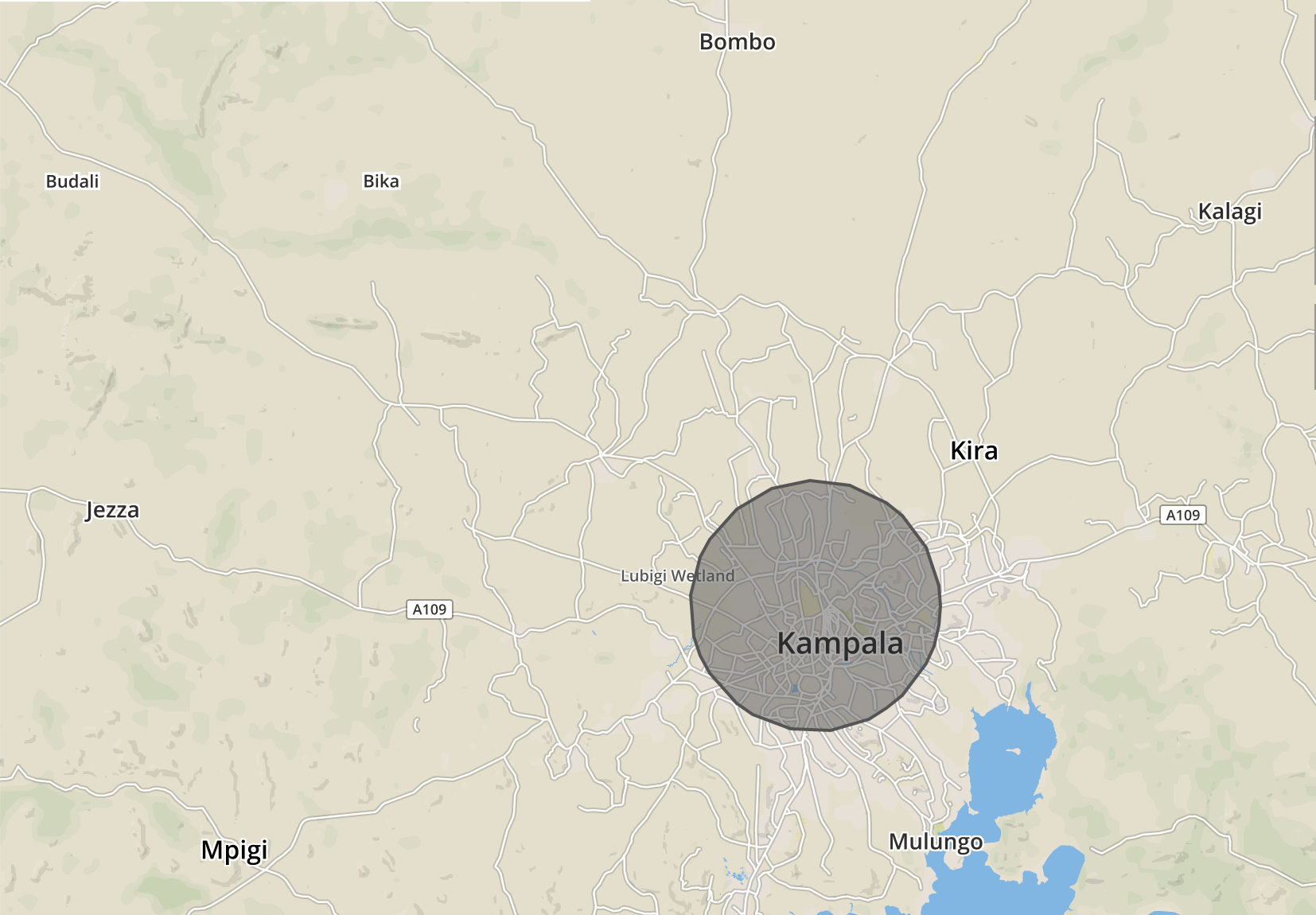}
    \hspace{0.3em}
    \includegraphics[width=0.23\textwidth]{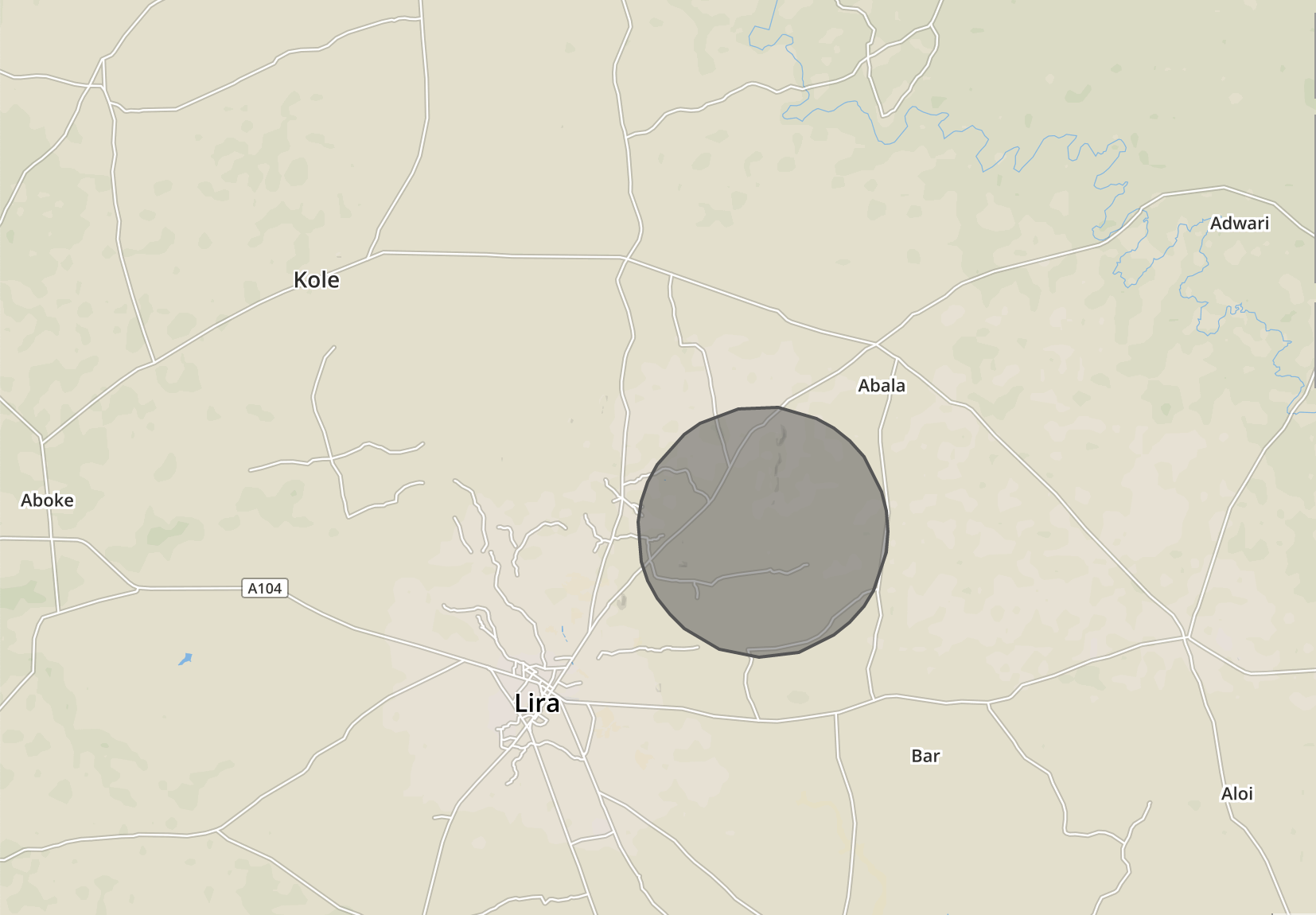}
    \caption{\textbf{Left}: A region with high wealth level and high truck numbers. A region with low wealth level and low truck numbers.}
    \label{fig:truckregions}
\end{figure}

\begin{figure}[!h]
    \centering
    \includegraphics[width=0.23\textwidth]{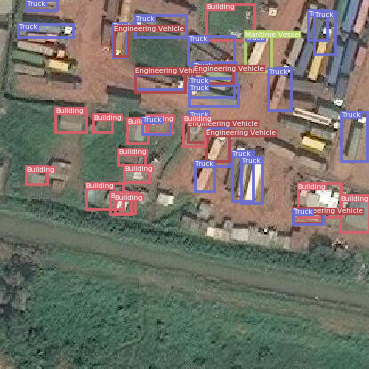}
    \includegraphics[width=0.23\textwidth]{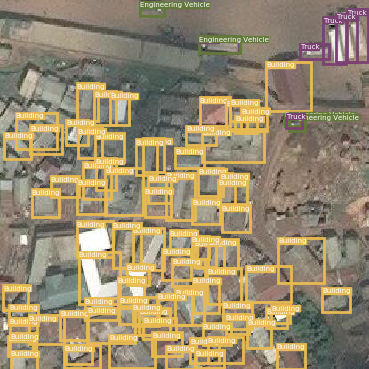}
    \hspace{0.3em}
    \includegraphics[width=0.23\textwidth]{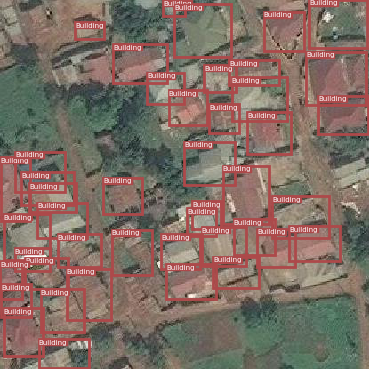}
    \includegraphics[width=0.23\textwidth]{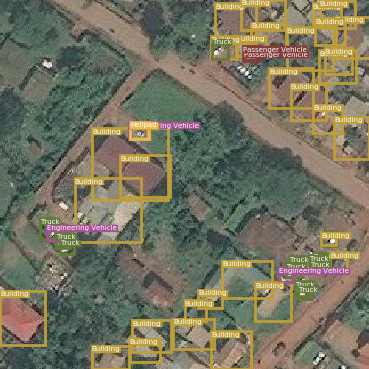}
    \includegraphics[width=0.23\textwidth]{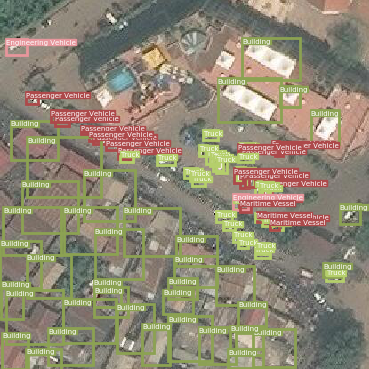}
    \includegraphics[width=0.23\textwidth]{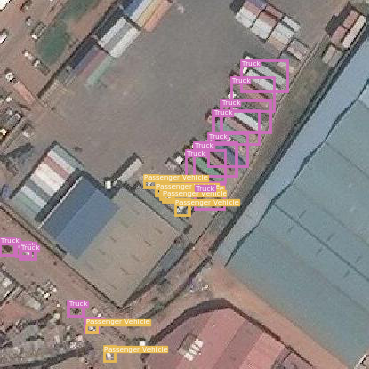}    
    \caption{Additional detection results from Uganda. Zoom-in is recommended to visualize the bounding boxes and labels. The detector can reliably detect buildings and trucks with reasonable overlap. However, it misses some of the small cars. Detecting small cars is a very challenging task due to very small number of representative pixels
    .}
    \label{fig:additionaldetections1}
\end{figure}
\begin{table}[!h]
\centering
\resizebox{0.69\columnwidth}{!}{
\begin{tabular}{@{}cll@{}}
\toprule
\textbf{Class} & \multicolumn{1}{c}{\textbf{ AP}} & \multicolumn{1}{c}{\textbf{ AR}} \\ \midrule
Aircraft Hangar & 0.000 & 0.000 \\
Barge & 0.010 & 0.040 \\
Building & 0.593 & 0.787 \\
Bus & 0.144 & 0.401 \\
Cargo Truck &  0.002 &  0.014 \\
Cargo/container Car & 0.272 & 0.519 \\
Cement Mixer & 0.002 & 0.012 \\
Construction Site & 0.015 & 0.067 \\
Container Crane & 0.015 & 0.040 \\
Container Ship & 0.218 & 0.304 \\
Crane Truck & 0.000 & 0.000 \\
Damaged/demolished Building & 0.000 & 0.000 \\
Dump Truck & 0.014 & 0.072 \\
Engineering Vehicle & 0.000 & 0.000 \\
Excavator & 0.250 & 0.462 \\
Facility & 0.000 & 0.000 \\
Ferry & 0.003 & 0.018 \\
Fishing Vessel & 0.002 & 0.010 \\
Fixed-Wing Aircraft & 0.003 & 0.002 \\
Flat Car & 0.018 & 0.018 \\
Front Loader/Bulldozer & 0.026 & 0.065 \\
Ground Grader & 0.000 & 0.000 \\
Haul Truck & 0.251 & 0.472 \\
Helicopter & 0.001 & 0.001 \\
Helipad & 0.050 & 0.050 \\
Hut/Tent & 0.000 & 0.000 \\
Locomotive & 0.000 & 0.000 \\
Maritime Vessel & 0.088 & 0.190 \\
Mobile Crane & 0.038 & 0.101 \\
Motorboat & 0.023 & 0.061 \\
Oil Tanker & 0.000 & 0.000 \\
Passenger Vehicle & 0.000 & 0.000 \\
Passenger Car & 0.486 & 0.759 \\
Passenger/Cargo Plane &  0.402 &  0.626 \\
Pickup Truck & 0.001 & 0.002 \\
Pylon & 0.337 & 0.589 \\
Railway Vehicle & 0.00 & 0.00 \\
Reach Stacker & 0.027 & 0.083 \\
Sailboat & 0.164 & 0.445 \\
Shed & 0.026 & 0.027 \\
Shipping Container & 0.002 & 0.018 \\
Shipping Container Lot & 0.185 & 0.398 \\
Small Aircraft & 0.109 & 0.288 \\
Small Car & 0.540 & 0.917 \\
Storage Tank & 0.000 & 0.000 \\
Straddle Carrier & 0.059 & 0.157 \\
Tank Car & 0.000 & 0.000 \\
Tower & 0.000 & 0.000 \\
Tower Crane & 0.013 & 0.033 \\
Tractor & 0.000 & 0.000 \\
Trailer & 0.046 &  0.237 \\
Truck & 0.189 & 0.637 \\
Truck Tractor & 0.000 & 0.004 \\
Truck Tractor w/ Box Trailer & 0.195 & 0.552 \\
Truck Tractor w/ Flatbed Trailer & 0.012 & 0.038 \\
Truck Tractor w/ Liquid Tank & 0.001 & 0.023 \\
Tugboat & 0.004 & 0.027 \\
Utility Truck & 0.000 & 0.000 \\
Vehicle Lot & 0.033 & 0.098 \\
Yacht & 0.064 & 0.162 \\ \bottomrule
\textbf{Total} & \textbf{0.082} & \textbf{0.163} \\ \bottomrule
\end{tabular}}
\caption{The AP and AR scores for child classes.}
\end{table}
Here we provide additional analysis to understand the effects of various features on the output of the model. Similar to Figure~\ref{fig:shap} (right) we plot (Figure~\ref{fig:shap1}) the SHAP value of a feature vs. the value of that feature for all the examples in the dataset.

Figure~\ref{fig:shap1} represents the change in predicted poverty score as the feature value under consideration changes and also reveals the interaction between that feature and another feature. Top left figure shows that regions with high \#\textit{Trucks} also have high \#\textit{Passenger Vehicles}. From top right figure, we find that higher \#\textit{Railway Vehicles} pushes the output to a higher value and low \#\textit{Railway Vehicles} has a negative impact on the output, thereby lowering the predicted value. We also observe that regions with high \#\textit{Railway Vehicles} also have high \#\textit{Passenger Vehicles}. On the other, we find (bottom left and bottom right figures) that \textit{Buildings} and \textit{Engineering Vehicles} do not tend to show much impact on the prediction value as their value increases.

Figure~\ref{fig:truckregions} compares an example region (left) with low poverty level and high \#\textit{Trucks} against a region with high poverty level and low \#\textit{Trucks}. We find that regions with high truck numbers have better road network and transportation connectivity to nearby regions, thereby resulting in better wealth index in those regions as transportation and connectivity play a vital role in the economic growth of a region.